\documentclass[journal]{IEEEtran}

\pdfoutput=1

\usepackage[utf8]{inputenc} 
\usepackage[T1]{fontenc}    
\usepackage{booktabs}       
\usepackage{amsfonts}       
\usepackage{nicefrac}       
\usepackage{microtype}      
\usepackage{lipsum}
\usepackage{balance}
\usepackage{cite}

\newcommand{\nop}[1]{}

\newcommand{\RNum}[1]{\uppercase\expandafter{\romannumeral #1\relax}}
\usepackage{color}
\usepackage{mathtools}
\usepackage{amsmath} 
\usepackage{graphicx} 
\usepackage{subfigure}
\usepackage{epstopdf}
\usepackage{multirow}
\usepackage{cleveref}
\usepackage[noend]{algpseudocode}
\usepackage{algorithmicx,algorithm}
\crefname{section}{§}{§§}
\Crefname{section}{§}{§§}

\begin{document}

\title{FedNILM: Applying Federated Learning to\\NILM Applications at the Edge}

\author{Yu~Zhang,
		Guoming~Tang,
		Qianyi~Huang,
		Yi~Wang,
		Xudong~Wang,
		Jiadong~Lou
		
\thanks{Y. Zhang, G. Tang, Q. Huang and Y. Wang are with the Peng Cheng Laboratory, Shenzhen, China. X. Wang and J. Lou are with the Chinese University of Hong Kong, Shenzhen, China.}
\thanks{Corresponding author: G. Tang (tanggm@pcl.ac.cn).}
}

\maketitle

\begin{abstract}
Non-intrusive load monitoring (NILM) helps disaggregate the household's main electricity consumption to energy usages of individual appliances, thus greatly cutting down the cost in fine-grained household load monitoring. To address the arisen privacy concern in NILM applications, federated learning (FL) could be leveraged for NILM model training and sharing. When applying the FL paradigm in real-world NILM applications, however, we are faced with the challenges of edge resource restriction, edge model personalization and edge training data scarcity.

In this paper we present FedNILM, a practical FL paradigm for NILM applications at the edge client. Specifically, FedNILM is designed to deliver privacy-preserving and personalized NILM services to large-scale edge clients, by leveraging i) secure data aggregation through federated learning, ii) efficient cloud model compression via filter pruning and multi-task learning, and iii) personalized edge model building with unsupervised transfer learning. Our experiments on real-world energy data show that, FedNILM is able to achieve personalized energy disaggregation with the state-of-the-art accuracy, while ensuring privacy preserving at the edge client. 
\end{abstract}


\section{Introduction}

\IEEEPARstart{N}{on-intrusive} load monitoring, also known as energy disaggregation, was first proposed by Hart in 1992~\cite{1992hart}. It is a single-channel blind source separation (BSS) problem that aims to decompose the aggregated power readings of a household into appliance-wise power consumption. One of the major purposes of NILM is to help reduce household energy consumption efficiently. Evidences have shown that such feedback of itemized information could encourage householders to use energy in a more sustainable way, achieving about $15\%$ energy saving~\cite{2008Feedback, ehrhardt2010advanced}. Besides, NILM can be leveraged to evaluate conservation programs, improve the quality of load forecasting, and provide references for power grid management~\cite{2011Disaggregated}. For example, with real-time NILM applications, utility companies could suggest switching operations on particular appliances (e.g., air conditioners) for load shifting in peak power hours~\cite{ehrhardt2010advanced}.

Although proposed for decades, the NILM problem has not been addressed completely, and traditional solutions referring to man-made appliance signatures have run into bottlenecks~\cite{beckel2014eco, zhong2014interleaved}. Most recently, it shows that the deep neural network (DNN) based approaches could greatly improve the performance of NILM~\cite{neuralnilm, zhang2016sequencetopoint, 2019Subtask, 2019tree}, as neural networks are able to automatically learn appliance features (or \emph{appliance signatures}), either obvious or latent ones. Various neural network architectures have been proposed for NILM, including denoising auto-encoder~\cite{neuralnilm}, recurrent neural networks~\cite{2019rnn}, and GAN~\cite{2020gan}, etc. Among those DNN based methods, the \emph{Seq2Point} model~\cite{zhang2016sequencetopoint}, a one-dimensional CNN based auto-encoder architecture, is the current state-of-the-art model for energy disaggregation. 

Arguably, the DNN based NILM models largely rely on sufficient and diverse training data, whereas realistic datasets often exist in the form of isolated islands. Although there are plenty of meter data in different buildings, it is almost impossible to transmit or integrate these local user data into a centralized storage, due to limits in communication bandwidth and legislation in user privacy and data security. Actually, in the last few years, the emphasis on data security and user privacy has become a global issue. For example, in the United States, China, and the European Union, relevant regulations have been enforced to protect data security and privacy~\cite{GDPR, chinacyberpower}, making it high-risk to gather massive user energy consumption data. In consequence, it is impractical to train powerful NILM models with existing paradigms.

\subsection{Motivations and Challenges}

To address such problems, federated learning (FL) has recently emerged as a promising paradigm. In a canonical FL system, user data is kept on client devices and the NILM model training is realized by i) local model updates with users' own data and ii) cloud model fusion with all users' models. In this way, client data remains locally and separate clients are integrated to build a up-to-date cloud model, thus allowing local models to collectively reap the benefits of cloud model trained from rich data~\cite{mcmahan2017communication}. It seems natural to apply the FL paradigm to the NILM problem for privacy preserving, which was theoretically verified in~\cite{wang2021federated} recently. When dealing with real-world NILM applications at local houses, however, we are still faced with the following major challenges.

\textbf{Challenge 1:} One critical limitation of adopting the typical FL system to NILM applications is the constrained resource in local homes (or \emph{edge clients}), in terms of computation power, communication bandwidth, memory and storage size, etc. In reality, usually low-end devices (or \emph{edge devices}) appear in local homes, rather than those powerful CPU/GPU servers. Hence, as the NILM models usually take large memory and computation overheads, network training on such resource-constrained edge devices could take prohibitively long time or even be terminated immediately. Since we cannot perform the model fusion at cloud without local updates, the resource lack at the edge client literally hinders the FL paradigm from applying in NILM applications.

\textbf{Challenge 2:} The second one relates to the personalization of NILM models. Although recent DNN based approaches are promising for NILM, it is not clear whether existing models are transferable among the diverse edge clients, since most of the models are trained on public datasets (e.g., REDD~\cite{2011REDD} and UK-DALE~\cite{ukdale2014}). In other words, even with satisfied performance on the common training datasets (or \emph{source domain}), these models may perform badly in a testing house (or \emph{target domain}), owing to the distribution difference between training and testing data~\cite{2019transfernilm}. Generally speaking, different households usually have different appliances as well as energy usage patterns, making those models hardly capture edge clients' heterogeneity and resulting in poor scalability in practice~\cite{2019transfernilm}.

\textbf{Challenge 3:} The third one is on training data scarcity at edge clients. For NILM applications, obtaining unlabelled testing data (i.e., mains power data) is quite easy, as current utility smart meters report the whole-home energy consumption periodically. To acquire labelled training data (i.e., power consumption data of individual appliances), however, is extremely expensive if not impossible. Although there are various power sensing devices and we could equip each appliance with a power sensor, this would incur enormous installation and maintenance costs, and thus is unscalable across large-scale households. As current DNN based NILM models are generally trained from large amounts of labelled data~\cite{2019transfernilm}, the scarcity of training data further limits the scalability of NILM systems and applications.

\subsection{Our Ideas and Contributions}

To tackle the resource limitation at the edge client ({Challenge~1}), our idea is to compress the complex NILM model to a simpler one that the edge device can afford. More specifically, the cloud NILM model under the FL paradigm could be pruned before being circulated to edge clients, and thus the edge devices would be able to implement local computation based on the compressed global state. Besides, a compressed NILM model is also desirable for cloud-edge communications. Then, to build personalized NILM models for diverse edge clients ({Challenge~2}), particularly under the condition of training data scarcity ({Challenge~3}), we propose to leverage the unsupervised transfer learning. By aligning the feature map distributions between the source and target domains, the unsupervised transfer learning techniques (e.g., CORAL in Sec.~\ref{subsec:coral}) could be incorporated with existing NILM modelling process, and properly address the domain shift issue for NILM model personalization.  

In this work we present FedNILM, a practical FL paradigm for real-world NILM applications at the edge. FedNILM aims to provide privacy-preserving and personalized energy disaggregation for the large-scale edge clients, aided by the cutting-edge DNN model compressing and transfer learning techniques at resource constrained edge devices. This paper shows that, the adapted FL framework can be highly preferable to be applied in tackling NILM problems, which is expected to promote NILM applications and make it practical in real-world implementations. Our major contributions can be summarized as follows.

\begin{itemize}
    \item We propose FedNILM, a FL paradigm designed for real-world NILM applications. Aided by the FL paradigm and with practical considerations, FedNILM is expected to provide scalable NILM services with the state-of-the-art accuracy across large-scale households while retaining data privacy for edge clients.
    \item We adopt cloud model compression for edge adoption in FedNILM. By exploring how compression techniques influence accuracy and overhead of the NILM model, we are able to effectively cut down the computation cost at the edge while retaining satisfied performance.
    \item We incorporate unsupervised transfer learning with FedNILM for client model personalization. By introducing the CORAL loss into the state-of-the-art NILM model, we manage to realize local transfer learning without relying on labelled training data at the edge client.
    \item We make extensive evaluations to validate the performance of FedNILM. The results demonstrate that, aided by the model compression at cloud and model personalization at edge, FedNILM can provide a comparable performance to the state-of-the-art without compromising the user privacy.
\end{itemize}

The rest of the paper is organized as follows. In Sec.~II, we give the background knowledge and related literature. Sec.~III briefs the state-of-the-art NILM model and the federated learning rationale. Then, we present the design of FedNILM in Sec.~IV, including the paradigm overview and workflow. Two key operations of FedNILM, cloud model compression and edge model personalization are presented in Sec.~V and Sec.~VI, respectively. The experimental evaluations are performed in Sec.~VII. The paper is concluded in Sec.~VIII.

\section{Background and Related Work}

Recently, there are growing interests on deploying NILM applications and systems by energy service providers, energy aggregators and distribution system operators~\cite{2020gan}. With the urgent request of privacy preserving, the FL paradigm has been exploited~\cite{wang2021federated}, where the NILM model/service is deployed/delivered across the edge clients for energy disaggregation. In real-world implementations, however, some challenges could arise.

\subsection{NILM Model Compression}

The first challenge is how to perform DNN-based NILM model inference at edge clients where only resource constrained devices could be installed. Deploying compressed NILM models on edge devices might be a promising approach. It has been demonstrated that model compression may not do harm to the NILM performance while significantly reducing the computational overhead~\cite{2020edgeml, 2020edgenilm}. 

There are various algorithms for compressing or pruning neural networks. In~\cite{2020edgenilm}, the authors leverage filter pruning~\cite{li2016filterpruning} and tensor decomposition~\cite{lebedev2014tensor} methods to compress the convolutional layers, where the former refers to sparsify the neural network by removing less important parameters and the later is to perform a low-rank decomposition of the learnt filter matrix. Surprisingly, experimental results in~\cite{blalock2020state, 2020edgenilm} show that compressing of neural networks might bring some performance gain due to better generalization in test cases.

Although the compressed NILM models could be deployed at the edge client for inference, the training of personal NILM models still needs to be conducted on the powerful GPU/CPU servers at the cloud end. Thus, one client needs to upload his/her energy data to the cloud for personal NILM model training, and the data privacy concern still exists.

\subsection{NILM Model Transfer}

The other challenge is on the degradation of the cloud model at the edge client. When applying the FL paradigm for NILM, the same model trained at the cloud would be delivered to various client ends. When the distributions between the client data and the cloud data are different, which is the most likely case, we would see more or less performance degradation from the NILM model at the client.

To this end, the transferability issue of NILM models has been preliminarily explored. In~\cite{2019transfernilm}, the authors investigate two transfer learning schemes, i.e., the appliance transfer learning (ATL) and cross domain transfer learning (CTL). \nop{The ATL refers to train an original CNN model based on dataset from one particular appliance, and fine-tune the fully connected layers in this CNN model with training samples from other appliances, whereas the cross domain transfer learning aims to tweak the pre-trained model with datasets from target domains. }Both ATL and CTL freeze the convolutional layers and merely tune the fully connected layers during retraining. Recently, apart from the canonical CNN structure, the authors of~\cite{2020gan} develop the TrGAN-NILM, which is based on the generative adversarial networks (GANs), to automatically extract common feature representations between source and target domains through minimizing the statistical distance between different domains. 

Nevertheless, the aforementioned transfer learning techniques are in need of labelled training data on target domain. In other words, they cannot been directly applied in the situation where the target domain is unlabeled, which is quite common in real-world NILM applications. In consequence, for buildings that are newly added to the NILM service provider portfolio, obtaining labelled appliance meter data and performing transfer learning are prohibitively expensive and time consuming.\nop{ Therefore, an unsupervised transfer learning technique could be more practical, which does not need any labelled training data at local clients while still achieving model transfer.}

\bigskip

In this work, we design FedNILM, a practical FL paradigm for real-world NILM applications. Particularly, FedNILM tackles the computation limitation issue by incorporating model compression techniques, and addresses the model personalization problem with unlabelled target domain.

\begin{figure}[t]
\centering
\includegraphics[width=0.45\textwidth]{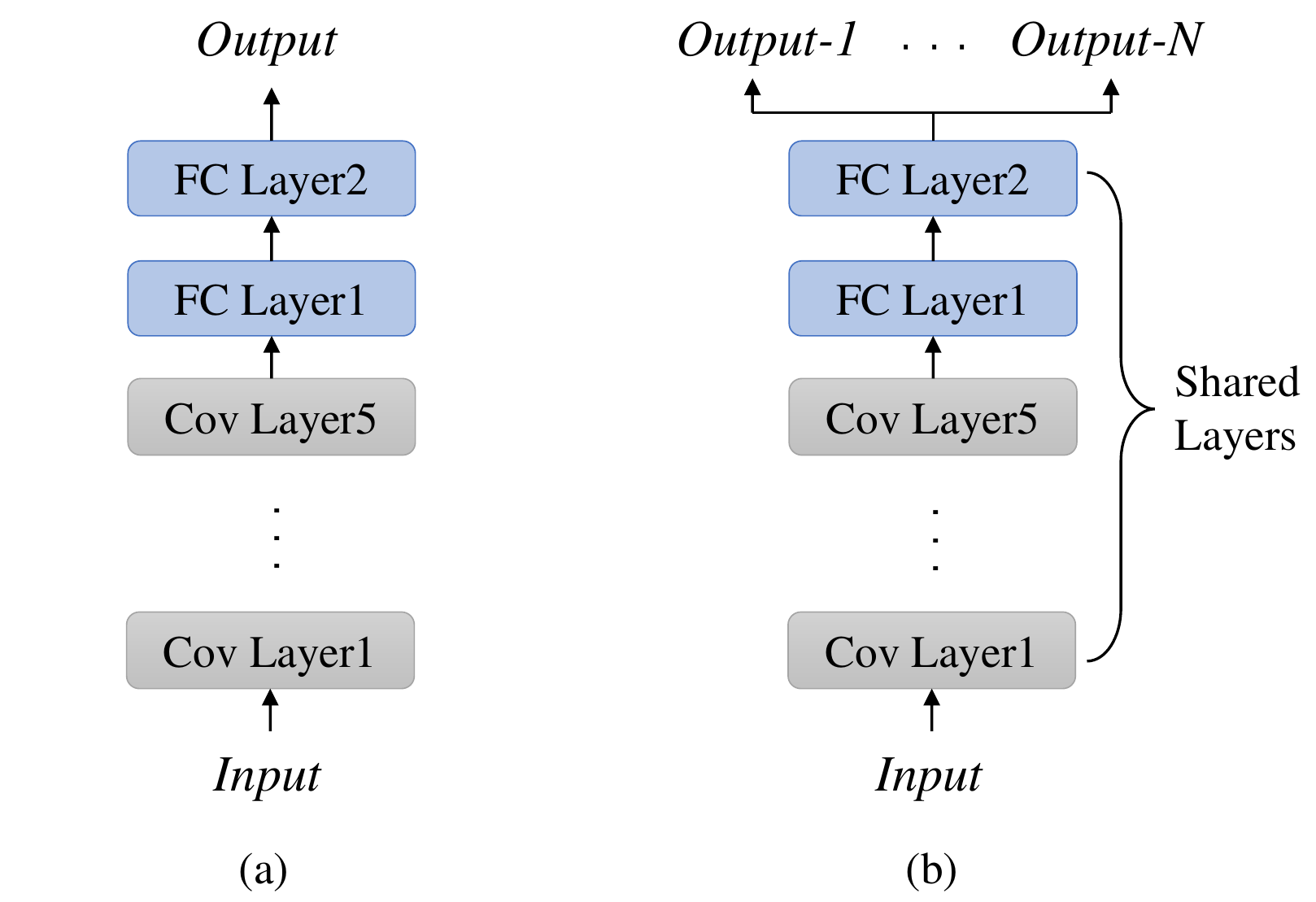}
\caption{(a) The DNN structure adopted by Seq2Point model. (b) The DNN structure for MTL-Seq2Point model, i.e., shared convolutional and fully connected layers for $N$ appliances.}
\label{fig:seq2point}
\end{figure}

\section{Preliminary}

\nop{
In this section, we give the definition of the NILM problem and introduce the adopted techniques in our federated learning paradigm for NILM applications.
}

\subsection{NILM Problem Definition}

The goal of NILM is to recover the energy consumption of individual appliances from the mains meter signals. Given the aggregate power readings from $T$ time periods, we can denote them by $\textbf{y} = (y_1, y_2, ..., y_T)$, where $y_t \in R_+$. Then, let $\textbf{x}^{(i)} = (x^{(i)}_1, x^{(i)}_2, ..., x^{(i)}_T)$ in which $x^{(i)}_t \in R_+$ denotes the power reading of the $i$-th appliance at time $t$. Therefore, at each time instant $t$, $y_t$ is assumed to be the sum of all $N$ appliances' power readings. Normally, we are only interested in the first $N'$ appliances used widely and consuming the most energy. Then, the power consumption of the remaining appliances can be represented as $\textbf{u} = (u_1, u_2, ..., u_T)$ and the aggregate power consumption could be represented as follows:
\begin{equation}
    y_t = \sum_{i=1}^{N'}{x^{(i)}_t} + u_t + \epsilon_t\label{(1)}
\end{equation}
where $\epsilon_t$ denotes a Gaussian noise.

\nop{
, e.g., the first $N'$ of the $N$ appliances ($N' \leq N$). Then,  Thus, Eqt.~(\ref{(1)}) can be modified as follows:
\begin{equation}
    y_t = \sum_{i=1}^{N'}{x^{(i)}_t} + u_t + \epsilon_t\label{(2)}
\end{equation}
Then, the purpose of NILM or ED is to estimate the power readings of $\textbf{x}^{(1)}, \textbf{x}^{(2)}, ..., \textbf{x}^{(N')}$ from $\textbf{y}$.

Essentially, NILM is a single-channel blind source separation (BSS) problem, which is unidentifiable and thus computationally intractable to solve, as we want to extract more than one source from a single observation. Thus far, most of NILM researches concentrate on improving the NILM accuracy and a variety of algorithms have been proposed to tackle this BSS problem, among which deep learning based NILM approaches showed performance superiority with almost 90\% error reduction compared to non-deep learning benchmarks, i.e., Combinatorial Optimisation (CO)~\cite{1992hart} and Factorial Hidden Markov Model (FHMM)~\cite{kim2011FHMM, 2011REDD}.}

\subsection{State-of-the-Art NILM Model}\label{subsec:seq2point}

We introduce the state-of-the-art model, i.e., sequence to point (Seq2Point)~\cite{zhang2016sequencetopoint}, recently developed for solving the NILM problem. The Seq2Point learning model maps a window of the mains signal readings to the midpoint point of the corresponding window of the target individual appliances. For each time instant $t$, given a fixed time window with size of $w$, the Seq2Point model uses the mains power signal sequence $\textbf{y}_{t:t+w-1} = [y_{t}, y_{t+1}, \cdots, y_{t+w-1}]$ as the input and the middle element $x^{(i)}_{t+w/2}$ in power readings of the target appliance $i$ as the output. In other words, instead of estimating the whole power signal sequence of the target appliance, the Seq2Point model merely predicts the middle signal element of the appliance in corresponding time window. 

\nop{The rationale behind Seq2Point is that, the appliance's power signal at any time should have a close relationship with the mains signals before and after that time instant.} 

Mathematically, for a target appliance $i$, it assumes that there exists a function $f^{(i)}: \mathbb{R}_+^{w} \rightarrow \mathbb{R}_+^{1}$, and the function gives the power estimation of appliance $i$ at time $t+ w/2$ by:
\begin{equation}
    f^{(i)}(\textbf{y}_{t:t+w-1}) = x^{(i)}_{t+w/2}
\end{equation}
Thus, the key task in Seq2Point is to learn the specific form of function $f^{(i)}$. Once obtained $f^{(i)}$, we are able to estimate the power signal of the target appliance $i$ with the aggregate (mains) signals, thus achieving energy disaggregation.

More specifically, to learn the parameters of $f^{(i)}$, Seq2Point employs the convolutional neural network (CNN) as the training structure, as illustrated in Fig.~\ref{fig:seq2point}(a). It was demonstrated that such a DNN structure could inherently learn the signatures of target appliances and shows superior performance than other models~\cite{neuralnilm}.

\nop{
Therefore, for an appliance $i$, the Seq2Point model can be represented by $f^i: \mathbb{R}_+^{w} \rightarrow \mathbb{R}_+^{1}$, and the power regression model of an individual appliance can be formulated as:
\begin{equation}
    f^i(\tilde{x}_{t}) = \tilde{p}_{t+w/2}^i
\end{equation}

For training network $f^i$, Seq2Point learning employs the convolutional neural network (CNN) as training structure, as it is demonstrated to be able to inherently learn the signatures of target appliances and thus show superior performance than recurrent neural network (RNN)~\cite{neuralnilm}. With time series sequences as input, all of the convolutional layers in Seq2Point model are 1-dimensional, and ReLU activations are exploited for all layers except the last one. The detailed structure of Seq2Point network is as follows (\textcolor{red}{better use a figure to illustrate the following texts.}):
\begin{enumerate}[(1)]
    \item Input (length determined by window size)
    \item 1D conv (filter size = 10, filter number = 30)
    \item 1D conv (filter size = 8, filter number = 30)
    \item 1D conv (filter size = 6, filter number = 40)
    \item 1D conv (filter size = 5, filter number = 50)
    \item 1D conv (filter size = 5, filter number = 50)
    \item Fully connected (N = 1024)
    \item Output(Fully connected, N = 1, activation = linear)
\end{enumerate}
}

\subsection{Federated Learning}\label{subsec:fl}

\nop{
The FL paradigm mainly consists of two procedures: homomorphic encrypted parameter sharing and federated aggregation~\cite{yangfederated}. 
}

At the beginning of FL, the server trains a cloud model based on the original dataset. When we adopt deep neural networks to learn both the cloud and client models, the learning objective of the global model could be formulated as:
\begin{equation}
    \mathop{\arg\min}_{\omega} f_s(\omega) = \frac{1}{n}\sum_{i=1}^n \mathcal{L}(x_i, y_i; \omega)
\end{equation}
where $\mathcal{L}(x_i, y_i; \omega)$ denotes the loss of prediction on server samples $\{x_i, y_i\}_{i=1}^n$ made with model parameters $\omega$, i.e., the weights and bias.

The server then sends the current global state $\omega$ to each of the clients, thus enabling each client to perform local computation based on the global state and its local dataset. Technically, the learning objective of client $u$ can be denoted as:
\begin{equation}
    \mathop{\arg\min}_{\omega^u} f_u(\omega^u) = \frac{1}{n^u}\sum_{i=1}^{n^u} \mathcal{L}(x_i^u, y_i^u; \omega^u)
\end{equation}
where $\{x_i^u, y_i^u\}_{i=1}^{n^u}$ are local samples with ${n^u}$ denoting their sizes. After the client model $f_u$ is trained, the local computation results, namely local model parameters $\omega^u$, would be uploaded to the server. For parameter transmission between cloud and client, homomorphic encryption is usually adopted to avoid information leakage~\cite{rivest1978data}. 

\nop{Since the encryption is not our main focus, we will show briefly the process of additive homomorphic encryption in parameter sharing. The additive encryption key pair created by the cloud server is denoted as $\left[\left[ \cdot \right]\right]$, and for any two real numbers $a$ and $b$, we have $\left[\left[ a \right]\right] + \left[\left[ b \right]\right] = \left[\left[ a + b \right]\right]$. After computing its local parameters $\omega^u$, each user would send the encrypted $\left[\left[ \omega^u \right]\right]$ to the server for aggregation. It has been shown that model with homomorphic encryption can still achieve a comparable accuracy compared with raw training model~\cite{yangfederated, 2019Quantifying}. Thus, the information transmitted for federated learning is minimal yet effective to improve a particular model. }

Then, with enough local updates, the cloud server performs federated aggregation~\cite{mcmahan2017communication} to align user models and obtain a new global state. Assuming that there are $K$ clients which are indexed by $k$ and each client locally updates its gradient to $\omega_{t+1}^k$ using its local data. The server then takes a weighted average of all these local models, which can be formulated as:
\begin{equation}
\label{federatedaveraging}
    \omega_{t+1} = \frac{1}{K}\sum_{k=1}^{K}\omega_{t+1}^k
\end{equation}
where $\omega_{t+1}$ denotes the updated global parameters. After adequate rounds of iterations and the global model has satisfactory generalization ability, the global model is distributed to the client for local deployment.

\nop{
\subsection{Multi-task Learning (\textcolor{blue}{move to latter section})}

Multi-task learning (MTL) refers to share representation, usually layers in neural network, between analogous tasks, and empowers the neural network to have better generalization ability~\cite{2017multitask}. Intuitively, leveraging excess information that comes from auxiliary tasks enables the MTL model to perform well in main task~\cite{caruana1997multitask}. The most frequently used approach of MTL in neural networks is hard parameter sharing where parameters of selected hidden layers are shared between all tasks, as illustrated in Fig.~\ref{fig1}.

As we can see, MTL model shares common layers and signatures for multiple analogous tasks. Unlike vanilla Seq2Point model which generates a separate model for each appliance, Seq2Point with multi-task learning  model train a single model for all appliances, significantly reducing the computation and memory overhead. To be more concrete, we leverage hard parameter sharing method in Seq2Point model, where the set of five convolutional layers and one fully connected layers are commonly shared for all appliances. Then after these shared layers, this model diverges to several task-specific layers for different appliances (see Fig.~\ref{fig2b}). Thus, the loss function of this Seq2Point model with MTL is the combination of losses from all appliances. 

\begin{figure}[t]
\centering
\includegraphics[scale = 0.4]{./fig/MTL.pdf}
\caption{Hard parameter sharing for multi-task learning in deep neural network.}
\label{fig1}
\end{figure}
}

\section{FedNILM Paradigm}

Modern DNN based NILM models cannot succeed without access to large amount of training data at the client side. However, this also causes severe concerns on user privacy and data security. To tackle these issues, we leverage the federated learning approach to the NILM problem and design FedNILM in this section. 

\nop{
... the cloud NILM model is compressed before being circulated to local clients, and thus the edge devices would be able to implement local computation based on the compressed global state and its own dataset. Previous work~\cite{2020edgenilm} shows that, compared with the original model, training a systematically compressed model could save approximately 95\% of run time and memory cost while retain a comparable accuracy. Therefore, compared to the NILM model with full parameter vector, a compressed version would be more desirable for both local updates and cloud transmission. The FL based approach with parameter pruning is thus promising in scaling the NILM model to practical multi-building scenarios.

Unsupervised transfer learning could be an appropriate approach to tackle both the personalization and unlabelled target domain issues at edge clients. More concretely, unsupervised transfer learning techniques usually work by aligning the feature map distributions, e.g., second-order statistics (covariances), of source and target domains~\cite{2016coral, 2016Deepcoral}, and thus dispense with the need of labelled data in the source domain. In particular, the unsupervised transfer learning approach, e.g., CORrelation ALignment (CORAL), could be directly incorporated into deep neural networks via introducing CORAL loss into the original loss function for both feature learning and domain adaptation.
}

\subsection{Overview}
\begin{figure}[t]
\centering
\includegraphics[width = \linewidth]{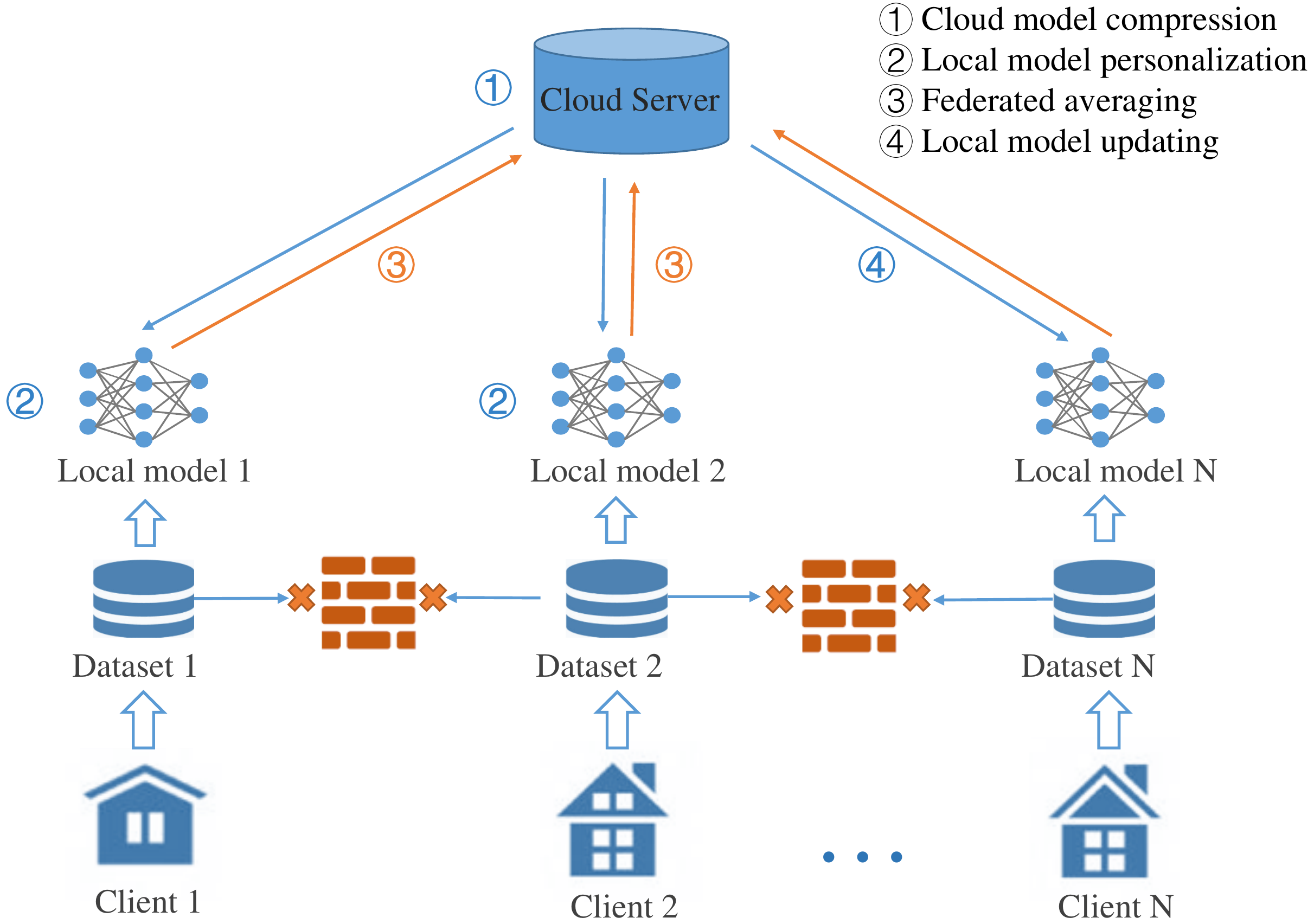}
\caption{Overview of the FedNILM framework.}
\label{fig2a}
\end{figure}

Without loss of generality, we consider one cloud server and multiple served clients at the edge, as illustrated in Fig.~\ref{fig2a}.
\begin{itemize}
\item At \textbf{cloud} side, the cloud server trains a Seq2Point model with readily-available open datasets, prunes it into a slim version, and shares it among the associated edge clients.
\item At \textbf{client} side, each takes over the pruned model from cloud, further tailors it through a local transforming process, and adopts the personalized model for individual appliance monitoring.
\end{itemize}

\subsection{Workflow}

As shown in Fig.~\ref{fig2a}, the workflow of FedNILM mainly consists of four steps. 
\begin{itemize}
\item \textbf{Step-1}: Based on the state-of-the-art NILM model (introduced in Sec.~\ref{subsec:seq2point}) and public datasets, the cloud server develops a compressed model with model pruning techniques. Refer to Sec.~\ref{subsec:cloudcompress} for the details of model pruning. Then, the compressed NILM model is distributed to all the associated edge clients.
\item \textbf{Step-2}: Based on the shared model from the cloud, each edge client further train a personalized NILM model with their own data at hand. An unsupervised transfer learning method is applied for the local model personalization. Refer to Sec.~\ref{subsec:clientpersonal} for the detailed transfer learning process.
\item \textbf{Step-3}: The parameters of personalized models at the edge are encrypted and uploaded to the cloud server, where the original cloud model is updated through the federated aggregation process (introduced in Sec.~\ref{subsec:fl}).
\item \textbf{Step-4}: The updated cloud model could be distributed to either new edge clients for personalized model building, or existing edge clients for continuous model refining with fresh local data.
\end{itemize}

Generally speaking, the complete FedNILM workflow includes the starting process (i.e., Step-1 and Step-2 in sequential) and a repeating process (i.e., Step-3 and Step-4 in iterative). Next we introduce the two key operations in the workflow, i.e., cloud model compression in Step-1 and client model personalization in Step-2, respectively.

\nop{
\begin{itemize}
    \item \textbf{Cloud model compression:} Owing to the limited resources on local devices, we perform model compression on step i) to extract a small cloud model from the original one. More specifically, we utilize filter pruning to prune the least important filters on convolutional layers in the initial cloud model and distribute this pruned version to local client devices, thus reducing the local computation and memory overhead and enabling local updates (see Sec.~\ref{sectioncompress} and Fig.~\ref{fig3}).
    \item \textbf{Local transfer learning:} Notice that in step ii) and step iv), as there exists large distribution divergence between server data (i.e., source domain) and user data (i.e., target domain), transfer learning is performed to align the second-order statistics of the source and target distributions~\cite{2016Deepcoral} and help train a more tailored model for each of local users (see Sec.~\ref{sectiontransfer} and Fig.~\ref{fig4}).
\end{itemize}
}


\section{Cloud Model Compression}\label{subsec:cloudcompress}

As we have mentioned in Sec.~\ref{subsec:seq2point}, the Seq2Point model trains a separate model for each individual appliance. That is to say, to monitor $n$ appliances for an edge client, we have to train and deploy $n$ appliance-specific models on the corresponding edge device. This could trigger tremendous resource demands and hardly be satisfied by the resource constrained edge devices. In this section, we show how model compression techniques could be leveraged in FedNILM.

\subsection{MTL-Seq2Point}

\nop{
It has been demonstrated that Seq2Point learning in ~\cite{zhang2016sequencetopoint} could reach the state-of-the-art performance with the ability to automatically extract instrumental features for NILM. Specifically, the Seq2Point architecture defines a neural network $f^i$ which maps the input mains meter sequence $\tilde{x}_t = (x_t, ..., x_{t+s-1})$ to the midpoint $\tilde{y}_t^i = y_{t+s/2}^i$ of output appliance's power sequence. In other words, the Seq2Point model forms a single prediction from the mains window to the midpoint of that appliance window. Therefore, the appliance power estimation model can be denoted as $f^i: \mathbb{R}_+^{s} \rightarrow \mathbb{R}_+^{1}$. Compared with sequence-to-sequence (Seq2Seq) learning~\cite{zhang2016sequencetopoint}, the main advantage of Seq2Point model is that there is a direct regression for each $y_t^i$, rather than an average of predictions for each input window~\cite{zhang2016sequencetopoint}. Therefore, we consider to leverage Seq2Point model as our basic structure, due to its architecture characteristics.
}

Multi-task learning (MTL) refers to share representation, usually layers in neural network, between analogous tasks, and empowers the neural network to have better generalization ability~\cite{2017multitask}. Intuitively, leveraging excess information that comes from auxiliary tasks enables the MTL model to perform well in main task~\cite{caruana1997multitask}. By leveraging MTL techniques in the Seq2Point modeling (we name the variant \emph{MTL-Seq2Point}), we are able to train one single model for all target appliances for NILM applications. This can significantly reduce the resource overhead at the edge device.

More specifically, we leverage hard parameter sharing method in Seq2Point model, where the set of five convolutional layers and one fully connected layers are commonly shared for all appliances. Then after these shared layers, this model diverges to several task-specific layers for different appliances. Refer to Fig.~\ref{fig:seq2point}(b) for the network structure of MTL-Seq2Point. In contrast with the original Seq2Point model, MTL-Seq2Point could help save large computation and memory overheads at the edge devices in our scenario.

\nop{
\begin{figure}[t]
\centering
\includegraphics[scale = 0.4]{./fig/MTL.pdf}
\caption{Hard parameter sharing for multi-task learning in deep neural network.}
\label{fig1}
\end{figure}

It has been demonstrated that the signatures learnt by different appliances are largely similar, including ``ON/OFF'' state changes, power levels and duration of specific activities~\cite{2019transfernilm}. Thus, it is intuitively feasible to share the same signature mappings across all appliances and merely train the last classification layers. The loss function for optimising the MTL model is combined with losses from all the appliances. 
}

\subsection{Pruned MTL-Seq2Point}\label{sectioncompress}

In addition to the multi-task learning, we also seek other effective techniques to further compress the NILM model in compatible with the less powerful edge devices. There has been some progress recently towards this direction, including weights pruning, filter pruning~\cite{li2016filterpruning}, neuron pruning and tensor decomposition~\cite{lebedev2014tensor}. As convolutional layers accounting for most of the computation cost in our NILM model~\cite{2020edgeml, 2020edgenilm}, we propose to leverage filter pruning in convolutional layers to build a ``slimmer'' MTL-Seq2Point model while retaining a comparable NILM performance. We name the model \emph{pruned MTL-Seq2Point} in this work.

\nop{
The primary goal of model compression is to extract a smaller yet comparably accurate model from the original model. There has been some progress recently towards this direction, including weights pruning, filter pruning~\cite{li2016filterpruning}, neuron pruning and tensor decomposition~\cite{lebedev2014tensor}, etc. As convolutional layers account for most of the computation cost in neural networks~\cite{2020edgeml, 2020edgenilm}, we propose to leverage filter pruning strategies in convolutional layers to reduce the computation cost of Seq2Point model. 
}

\begin{figure}[t]
\centering
\includegraphics[scale = 0.4]{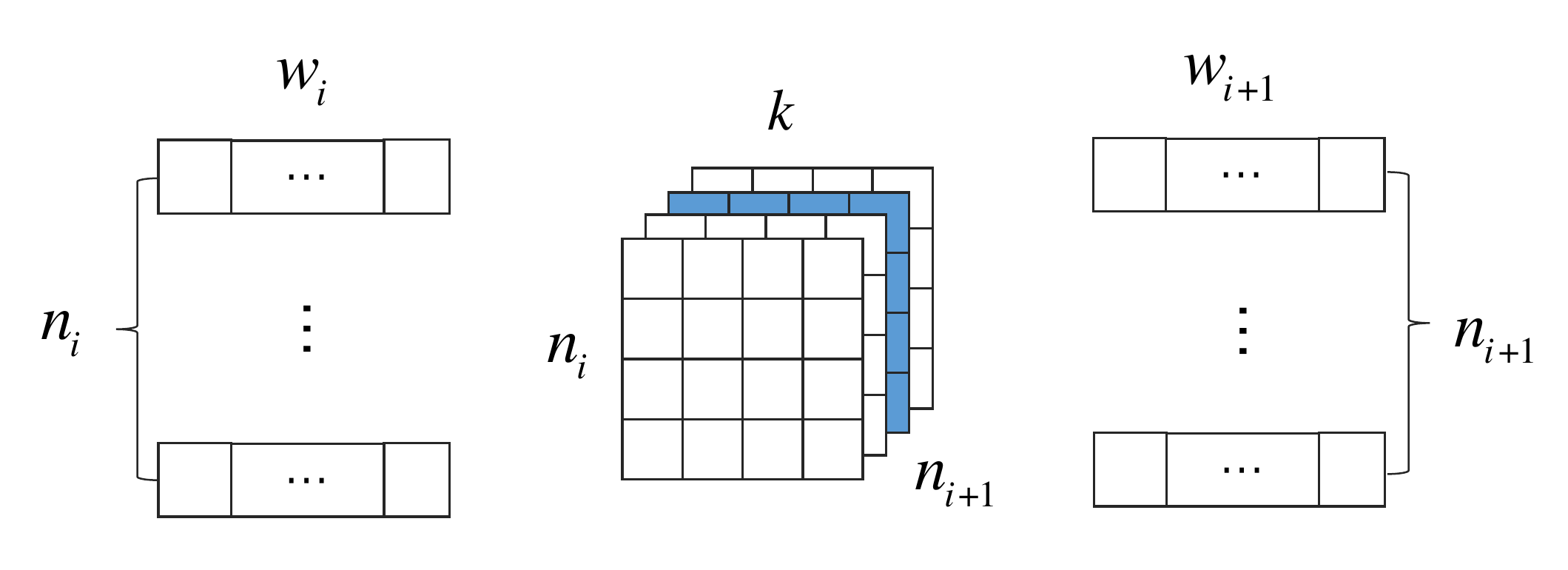}
\caption{Pruning a filter results in reduction of computation overhead.}
\label{fig3}
\end{figure}

Specifically, in the network structure of MTL-Seq2Point, let $n_i$ denote the number of input channels for the $i$-th convolutional layer and $w_i$ the window size of input. Given the kernel size $k$, we have 1D kernel $\mathcal{K} \in \mathbb{R}^{k \times 1}$ (e.g., 10 $\times$ 1 in the first layer). Supposing that the number of output channels in the $i$-th convolutional layer is $n_{i+1}$, the 3D filters matrix $\mathcal{F}_i \in \mathbb{R}^{n_{i+1} \times n_i \times k}$ could transform the input $\mathrm{X_i} \in \mathbb{R}^{n_i \times w_i}$ into the output $\mathrm{X_{i+1}} \in \mathbb{R}^{n_{i+1} \times w_{i+1}}$. Then, given the stride size $s$, we have $w_{i+1} = w_i-k+s$. Thus, the number of operations in the $i$-th convolutional layer is $n_{i+1} n_i k w_{i+1}$. By removing one of the $n_{i+1}$ filters in $\mathcal{F}_i$, we could eliminate $n_i k w_{i+1}$ operations, as illustrated by Fig.~\ref{fig3}. Moreover, pruning a filter also results in the removal of corresponding feature maps and kernels of the following layer, thus cutting down another $n_{i+1}kw_{i+2}$ operations. Hence, we can conclude that, by pruning $m$ filters of layer $i$, we could reduce the computation cost by $m/n_{i+1}$ at both the $i$-th and $(i+1)$-th layers.

Since not all trained filters are equally important, thus in order to minimize the performance drop, we choose and prune the less instrumental filters. More specifically, the relative importance of the filters are measured through calculating the sum of $\mathcal{L}_1$-norm~\cite{li2016filterpruning} or $\mathcal{L}_{2}$-norm~\cite{zhou2016filterprune, wen2016prune} on convolutional filters. As there are no noticeable differences between these two criteria in filter selection~\cite{li2016filterpruning}, we leverage the $\mathcal{L}_1$-norm to score the filters and prune $k$\% of them in each layer with the least $\mathcal{L}_1$-norm values. By increasing the pruning percentage of the network iteratively, we can also investigate the variation of model performance and thus find the optimal pruning percentage for our model. 

Overall, the procedure to obtain the best pruned MTL-Seq2Point model consists the following four steps: i) train a convergent MTL-Seq2Point model, ii) score corresponding convolutional filters based on the sum of $\mathcal{L}_1$-norm values, iii) prune the least important filters as per their scores, and iv) retrain the pruned MTL-Seq2Point model (for a certain iterations) to compensate for incurred performance degradation.

\section{Client Model Personalization} \label{subsec:clientpersonal}

When adopting the compressed cloud NILM model at the edge client, however, we are faced with the domain shift problem, i.e., the difference of distributions between the cloud and client data. Thus, the common model can perform very well upon the cloud dataset, while may be greatly degraded at different client ends. To this end, we leverage the transfer learning technique and build personalized models for different clients. Particularly, we adopt the unsupervised transfer learning to work with unlabelled client data, hence addressing the training data scarcity problem at the edge client.

\subsection{Transferability Analysis} \label{subsec:transfer_analysis}

Transfer learning works under the scenario where observations from the source domain (denoted by $D_s$) have a different distribution with those from the target domain (denoted by $D_t$). In our case, the source domain refers to the public dataset at cloud, while the target domain is the personal dataset at each client. There have been interests towards identifying the transferability of features in each layer of the DNN model, especially in Computer Vision~\cite{yosinski2014transferable}.

For our NILM model, to investigate the transferability of different model layers, we propose the following three-step procedure. 
\begin{enumerate}
    \item Train an initial NILM model (i.e., the pruned MTL-Seq2Point model) on $D_s$.
    \item Fix and fine-tune one of the multiple layers in the pre-trained model, and randomly initialize the parameters in the rest of layers.
    \item Retrain the model on $D_t$ and compare the prediction performance from the multiple transferred models.
\end{enumerate}

Note that the NILM model in our scenario is composed of five convolutional layers and two fully connected layers, as illustrated in Fig.~\ref{fig:seq2point}. Thus, through procedurally freezing or tweaking these layers, we have $(2 \times 7)$ different transfer learning models, which are then retrained and tested on target domain data. By wrapping up all the results, we have the following observations (refer to Sec.~\ref{subsec:localtransfer} for more details).

\begin{itemize}
    \item The convolutional layers of the model are good at extracting low-level and generic features. The transferable load features, such as the ON/OFF switching points, power level of appliances and typical usage durations, are insusceptible to the difference between source and target domains.
    \item The fully connected layers take responsibility for learning high-level features for specific appliances. Thus, when applying to a new edger client (target domain), they need to be further fine-tuned. The client model personalization can then be realized by largely referring to the transferred features from convolutional layers and substantially fine-tuning the fully connected layers for specific appliances.
\end{itemize}

These observations are coincide with those from~\cite{2019transfernilm} that, the features in lower levels of layers are highly transferable as lower layers tend to learn common and coarse information, whereas the features in higher layers are more tailored for specific tasks and thus are more personalized. Accordingly, in our case of transfer learning at the client side, we propose to freeze the convolutional layers in model transforming (i.e., keep their parameters fixed in back propagation) and merely update the weights on fully connected layers.

\subsection{Correlation Alignment}\label{subsec:coral}

\begin{figure}[t]
\centering
\includegraphics[scale = 0.35]{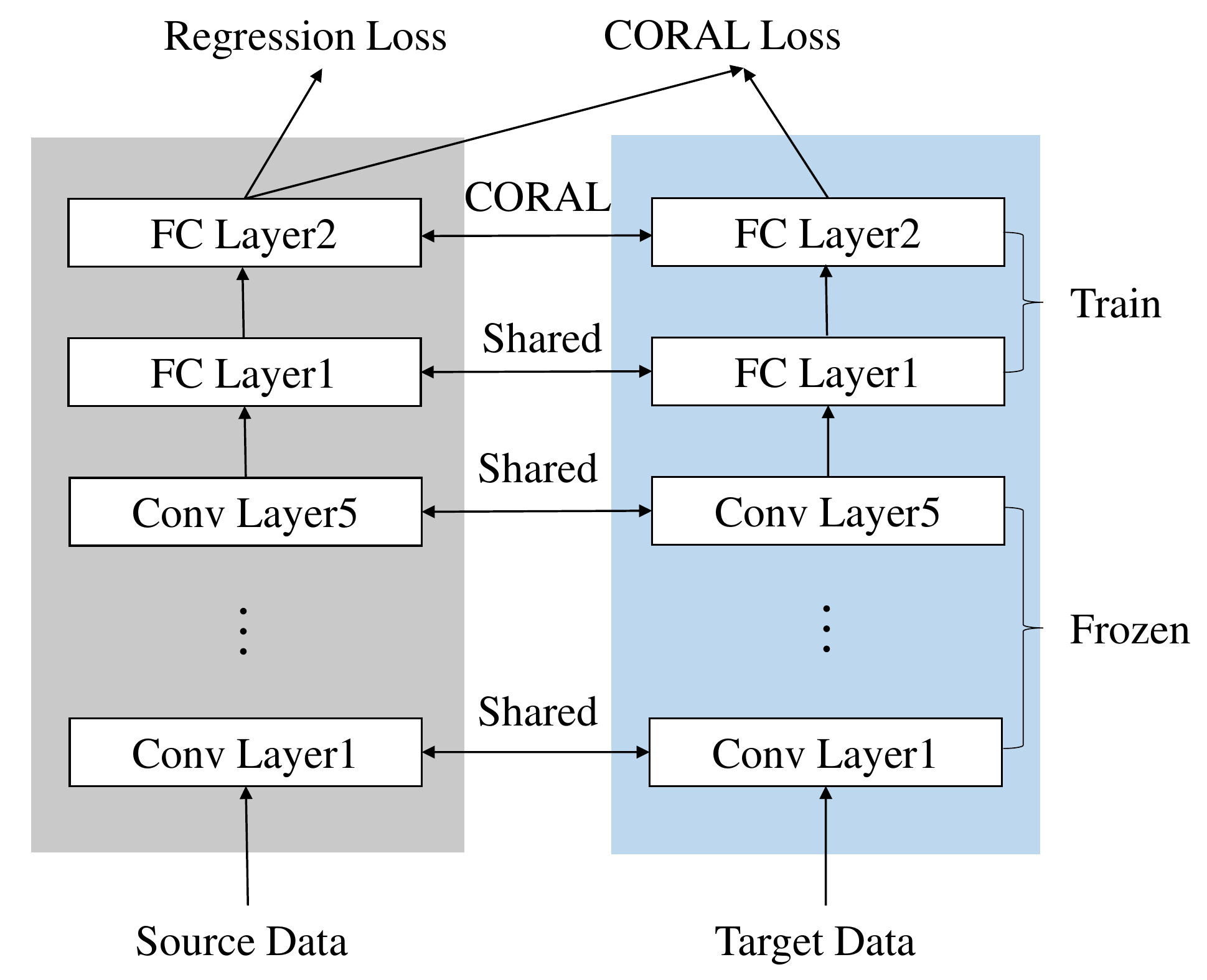}
\caption{The transfer learning process at the edge client.}
\label{fig4}
\end{figure}

Correlation alignment (CORAL) is an instrumental domain adaptation method, which tackles the domain shift via aligning the feature distributions of source and target samples~\cite{2016coral}. More specifically, deep CORAL aligns the source and target data distributions through learning a nonlinear transformation, i.e., a differentiable loss function (CORAL loss), between source and target layer activations~\cite{2016Deepcoral}. This nonlinear transformation is designed to minimize the distance between the second-order statistics (covariance) of the source and domain layer features. 

Given the source domain training data $D_s$ (with labels $L_s$) and target input data $D_t = {u_i}$, we aim to align the distribution difference in layer $I$ which generates the $\sigma$-dimensional deep layer features $x$ of input $D_s$ and $u$ of input $D_t$. With $x$ and $u$, we could compute the covariance matrices $C_s$ and $C_t$. Therefore, the CORAL loss is defined as the distance between the second-order statistics, namely covariance, of the source and target features:
\begin{equation}
    \mathcal{L}_{CORAL} = \frac{1}{4\sigma^2}\|C_s - C_t\|^2_F
\end{equation}
where $\|\cdot\|^2_F$ denotes the squared matrix Frobenius norm. Let $\lambda$ denote the trade-off parameter and $\mathcal{L}_{R}$ the regression loss of the edge model. The loss in the client model training can be designed as:
\begin{equation}
    \mathcal{L}_{EDGE} = \mathcal{L}_{R} + \lambda\cdot\mathcal{L}_{CORAL}
\end{equation}
The above loss serves as a constraint and regulates the distance between source and target domains during the fine-tuning process. By jointly optimizing the regression loss and CORAL loss, we can obtain a personalized client model with both generic signatures pre-trained on the source domain and specific features working well on the target domain. 

Fig.~\ref{fig4} shows the detailed model structure tailored for the unsupervised transfer learning process. In this way, the local devices are able to reap the shared benefits of common features pre-trained on a large generic dataset while retaining the speciality fine-tuned on client data. 

\section{Experiments}

In this section, we use real-world energy datasets to evaluate the proposed FedNILM paradigm, including cloud model compression and client model personalization, respectively.

\subsection{Experimental Settings}

\subsubsection{Cloud and Edge Systems}
In our experiments, an AMAX server with four Tesla V100 GPUs serves as the cloud end. A low-end desktop (with Nvidia GTX 960M) and an edge device (Nvidia Jetson Nano\nop{ with a 1.43 FHz CPU and 4 GB of RAM}) serve as two different client ends. \nop{At each client end, we store the power consumption data of several houses in REDD, UK-DALE and REFIT datasets. }

\nop{Generally, we experiment our paradigm over four appliances: the fridge, dish washer, washing machine and microwave. We exploit these four target appliances for the following reasons. First, they collectively consume a significant proportion of household energy consumption. Second, they are commonly shared in the datasets and thus can be leveraged to test the model transferability. Third, they represent a range of different operation patterns. For example, a fridge is an appliance which always runs in the background while a washing machine is an interactive-based appliance.}

\subsubsection{Datasets}
Three benchmark datasets are used to evaluate the performance of FedNILM, including REFIT~\cite{refit2017}, UK-DALE~\cite{ukdale2014} and REDD~\cite{2011REDD}. All contain similar appliance categories, allowing the evaluation of model transferability. Also, they have been widely applied in previous NILM researches~\cite{neuralnilm, zhang2016sequencetopoint, 2019transfernilm}, and thus enabling performance comparison with the state-of-the-art solutions.

\nop{
\subsubsection{REFIT}

The REFIT dataset contains electrical consumption of 20 UK households from October 2013 to July 2015 with 8-second sampling intervals. In particular, we leverage the dataset from house 2 as public dataset to train cloud model, datasets from house 5, 6 as clients' private data to perform local computation.

\subsubsection{UK-DALE}

In UK-DALE, both mains and appliances power readings were recorded once every 6 seconds from November 2012 to January 2015. This dataset contains aggregate power consumption and measurements of $4-54$ appliances of five UK houses. For evaluation, we use the data of house 1 as the cloud training set, house 2 and 5 as the local training set. 

\subsubsection{REDD}

The REDD dataset contains power sequence data for six US houses, with 1 second recording frequency for mains meter and 3 seconds for $10-25$ types of appliances. The lengths of observations range from 3 days to 19 days. We exploit house 1 for cloud training, and house 2 and 3 for local updating.

All these datasets contain low frequency measurements ($\leq 1 Hz$), ranging from 1 Hz to 8 Hz. We thus down-sample the power sequences to 60 seconds as in previous literature~\cite{2019Subtask, 2020edgenilm}. Meanwhile, the input sequence length is a hyper-parameter in Seq2Point model and we choose to experiment over two different sequence lengths: 99 and 499. The training windows are preprocessed by substracting the mean values and dividing by the standard deviations.
}

\subsubsection{Performance Metrics}
Mean absolute error (MAE), signal aggregate error (SAE) and F1-score are used to evaluate the performance of FedNILM, all of which have been leveraged in prior NILM research~\cite{neuralnilm, zhang2016sequencetopoint, 2019Subtask, 2020edgenilm}. In particular, the former two metrics, i.e., MAE and SAE, aim to measure the performance of power consumption estimation, while the F1-score could reflect the performance of appliance ON/OFF states estimation.

\begin{table}[t]
\caption{MAE results from cloud model compression on REDD dataset (sequence length 499)}
\begin{tabular}{|c|cccc|cc|}
\hline
Model     & \begin{tabular}[c]{@{}c@{}}Washing \\ m.\end{tabular} & Fridge & \begin{tabular}[c]{@{}c@{}}Dish \\ w.\end{tabular} & \begin{tabular}[c]{@{}c@{}}Micro\\ wave\end{tabular} & \begin{tabular}[c]{@{}c@{}}Time\\ (s)\end{tabular} & \begin{tabular}[c]{@{}c@{}}Size\\ (MB)\end{tabular} \\ 
\hline
\hline
Seq2Point       & 20.25                                                 & 25.61  & 14.09                                              & 13.30                                                & 9.36                                               & 367.82                                              \\ \hline
30\% pruning & 18.03                                                 & 26.16  & 12.82                                              & 9.95                                                 & 4.78                                               & 180.05                                              \\
60\% pruning & 18.30                                                 & 25.75  & 16.91                                              & 10.65                                                & 3.06                                               & 58.80                                               \\
90\% pruning & 20.53                                                 & 44.57  & 48.36                                              & 13.76                                                & 2.33                                               & 3.69                                                \\ 
\hline
\hline
MTL-Seq2Point       & 18.74                                                 & 27.04  & 19.40                                              & 12.37                                                & 2.69                                               & 91.97                                               \\ \hline
30\% pruning & 18.55                                                 & 27.93  & 21.94                                              & 12.74                                                & 1.26                                               & 45.02                                               \\
60\% pruning & 19.18                                                 & 28.39  & 19.26                                              & 13.15                                                & 0.83                                               & 14.7                                                \\
90\% pruning & 19.39                                                 & 75.21  & 28.86                                              & 20.60                                                & 0.68                                               & 0.93                                                \\ \hline
\end{tabular}
\label{tbl:rlt_cloud_499}
\end{table}

\begin{table}[t]
\caption{MAE results from cloud model compression on REDD dataset (sequence length 99)}
\begin{tabular}{|c|cccc|cc|}
\hline
Model     & \begin{tabular}[c]{@{}c@{}}Washing \\ m.\end{tabular} & Fridge & \begin{tabular}[c]{@{}c@{}}Dish \\ w.\end{tabular} & \begin{tabular}[c]{@{}c@{}}Micro\\ wave\end{tabular} & \begin{tabular}[c]{@{}c@{}}Time\\ (s)\end{tabular} & \begin{tabular}[c]{@{}c@{}}Size\\ (MB)\end{tabular} \\ 
\hline
\hline
Seq2Point       & 14.97                                                 & 25.80  & 19.35                                              & 9.16                                                 & 3.13                                               & 55.32                                               \\ \hline
30\% pruning & 14.59                                                 & 23.77  & 11.97                                              & 9.86                                                 & 2.74                                               & 27.10                                               \\
60\% pruning & 14.73                                                 & 22.71  & 16.60                                              & 8.65                                                 & 2.67                                               & 8.87                                                \\
90\% pruning & 15.46                                                 & 30.70  & 32.23                                              & 10.52                                                & 2.55                                               & 0.58                                                \\ 
\hline
\hline
MTL-Seq2Point       & 17.62                                                 & 26.78  & 16.06                                              & 14.51                                                & 0.83                                               & 13.84                                               \\ \hline
30\% pruning & 17.26                                                 & 25.40  & 13.05                                              & 12.65                                                & 0.75                                               & 6.79                                                \\
60\% pruning & 18.67                                                 & 28.78  & 10.66                                              & 15.78                                                & 0.72                                               & 2.22                                                \\
90\% pruning & 19.59                                                 & 30.45  & 18.34                                              & 18.83                                                & 0.74                                               & 0.15                                                \\ \hline
\end{tabular}
\label{tbl:rlt_cloud_99}
\end{table}

\begin{figure*}[t]
\centering
\subfigure[MAE]{
\begin{minipage}[t]{0.31\linewidth}
\includegraphics[width=1\textwidth]{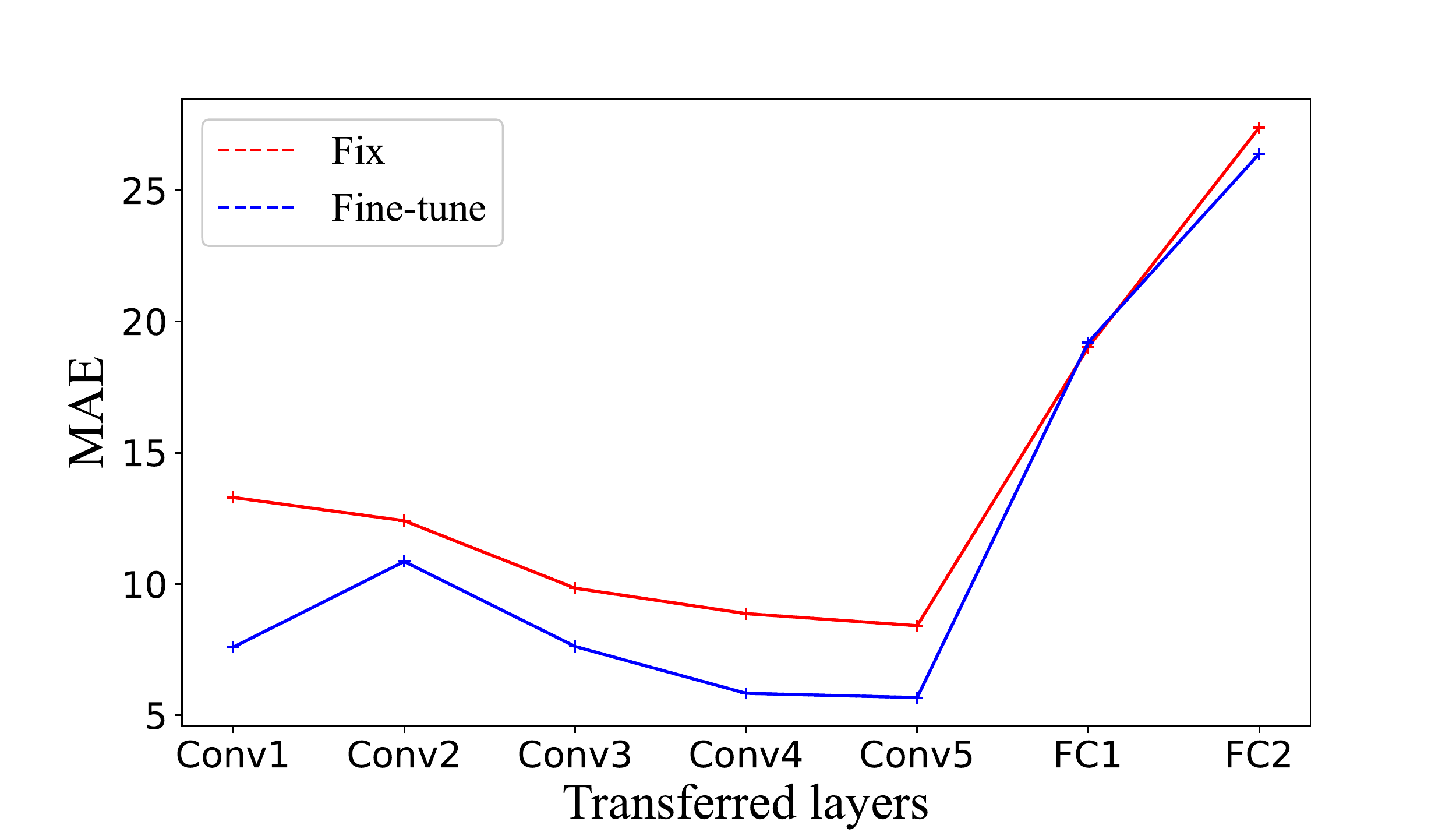}
\end{minipage}
}
\subfigure[SAE]{
\begin{minipage}[t]{0.31\linewidth}
\includegraphics[width=1\textwidth]{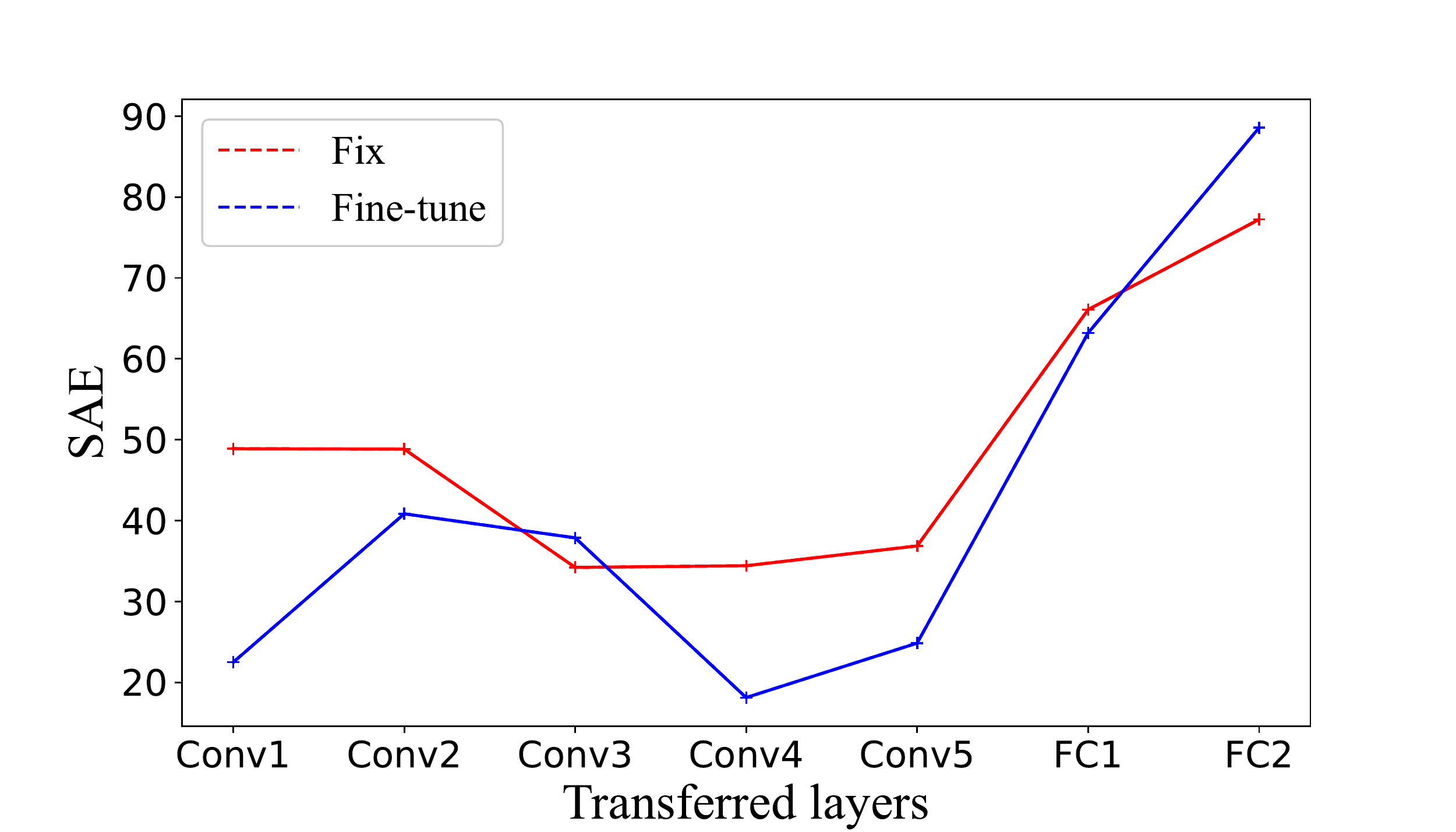}
\end{minipage}
}
\subfigure[F1-score]{
\begin{minipage}[t]{0.31\linewidth}
\includegraphics[width=1\textwidth]{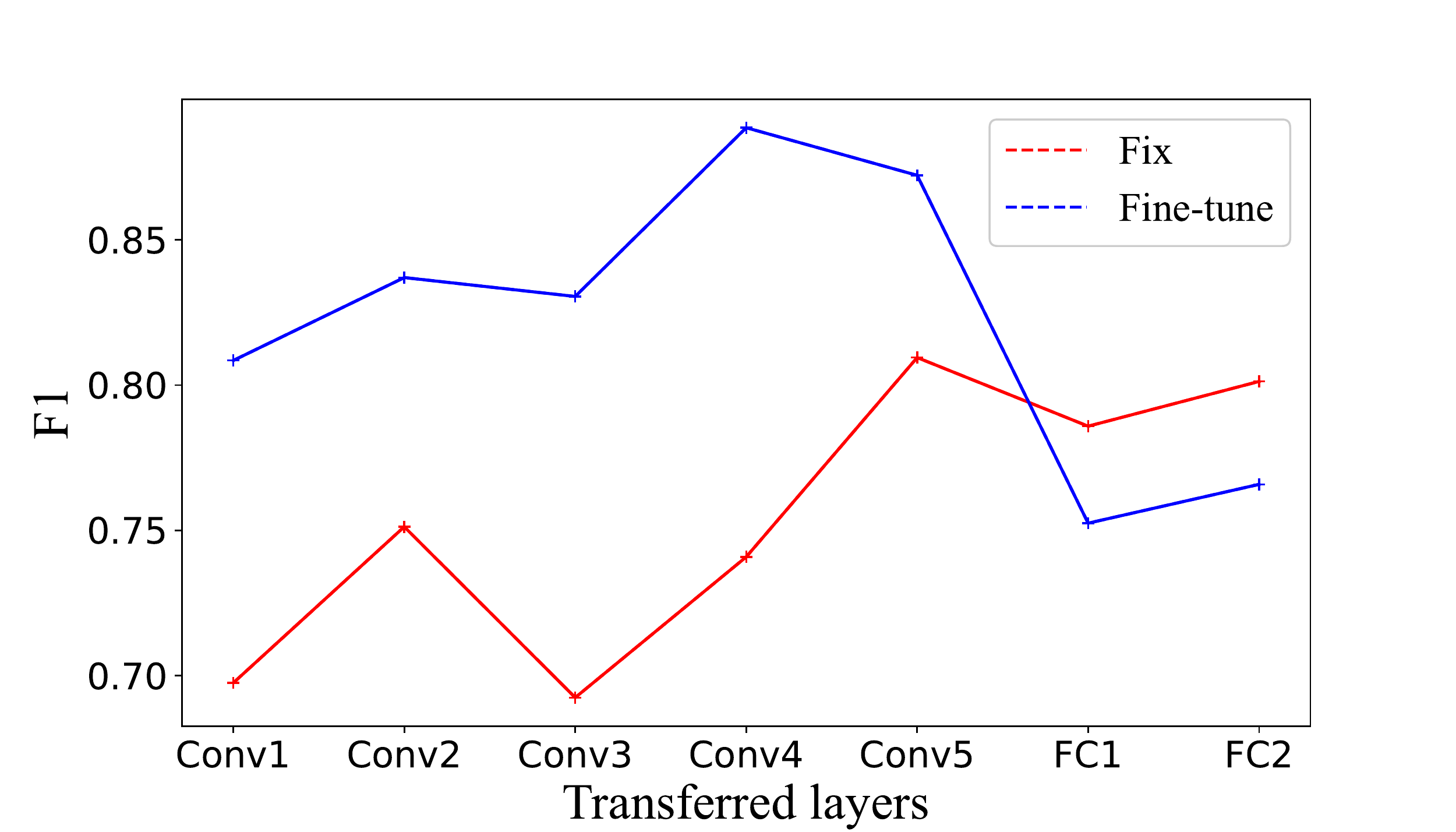}
\end{minipage}
}
\caption{(a) The MAE of transferred models (lower is better). (b) The SAE of transferred models (lower is better). (c) The F1-score of transferred models (higher is better). The x-axis indicates till which layer we transfer the parameters and initialize the rest. }
\label{fig5}
\end{figure*}

\subsection{Evaluations on Cloud Model Compression}\label{modelcompress}

At the cloud end, we train the Seq2Point model with/without pruning and MTL on the REDD dataset, and compare the NILM performance, run-time and memory overhead from different model adoptions. \nop{The reason why we employ MAE, not F1-score, to demonstrate the pruned model performance is that filter pruning might lead to more zeros in final output sequences due to limited parameters.}To show the impacts of model compression, we prune the parameters of Seq2Point and MTL-Seq2Point by 30\%, 60\% and 90\%, respectively. The results are summarized in Table~\ref{tbl:rlt_cloud_499} and Table~\ref{tbl:rlt_cloud_99}, for input sequence lengths of 499 and 99, respectively. \nop{ The S2P denotes the original Seq2Point model, and S2P(30\%), S2P(60\%) and S2P(90\%) denote its pruned version with different pruning percentage. Similarly, the MTL denotes the Seq2Point model with multi-task learning, and MTL(30\%), MTL(60\%), MTL(90\%) denote its pruned variants with different pruning proportion.}

Based on the experimental results in Table~\ref{tbl:rlt_cloud_499} and Table~\ref{tbl:rlt_cloud_99}, we observe that both multi-task learning and filter pruning could significantly reduce the model size along with decent inference time benefits, without drastically compromising NILM performance. In particular, for the sequence 499, pruning 60\% of filters on classical Seq2Point model can save approximately two-thirds of running time and reduce memory cost by up to 84\% while retaining comparable prediction accuracy. Meanwhile, the multi-task learning structure virtually takes one fourth of the original computation and memory overhead, as it simultaneously generates the disaggregation results of four selected appliances. The combination of filter pruning and multi-task learning techniques could help save nearly 92\% of inference time and 96\% of model space with slight (<10\%) performance degradation. 

For sequence length 99, the unoptimised model occupies $6\times$ more space and requires more inference time than the optimised model pruned 60\% of filters with similar performance. Same as we have found with the model of sequence length 499, for sequence length 99, the multi-task learning structure helps save nearly three fourths of running time and space, whereas filter pruning may not have much influence on inference time as the time required for these three pruned models with different pruning percentages are almost the same. Note that we merely prune the filters in convolutional layers and in a forward pass. The majority of computations takes place in fully connected layers, not convolutional layers, so the reduction of parameters in convolutional layers may not reduce much operations in inference procedure, thus explaining this phenomenon.

In conclusion, the general NILM performance decreases with the increasing of pruning percentage, and the employment of multi-task learning structure also contribute to the ED performance degradation as all the four appliances use the same set of signatures for prediction. However, we also observe that in some circumstances, model pruning might rather give rise to better NILM results, such as the pruned Seq2Point model performance versus the original Seq2Point model accuracy on distinguishing the washing machine, which can be explained by ``regularisation effect'' as removing the least important weights from a neural network might lead to better model generalisation. In general, we choose to build our general cloud model based on Seq2Point model with multi-task learning and prune 60\% of the filters in its convolutional layers.

\subsection{Evaluations on Client Model Personalization}
\label{subsec:localtransfer}

At the client end, we investigate the transferability of each layer of the NILM model, determine which layer's parameters to freeze or fine-tune during model transfer, and validate the necessity to perform local transfer learning.

\subsubsection{Layer Transferability of NILM Model}

\nop{
To begin with, the personalization and unlabelled target domain issue can be solve through unsupervised transfer learning techniques, i.e., CORrelation ALignment (CORAL) in this paper. The typical Deep CORAL approach set to apply the CORAL loss to the last dense layer~\cite{2016Deepcoral} and merely fine-tune the fully connected layers yet freeze all the convolutional layers~\cite{chen2020fedhealth} based on the usual assumption that convolutional layers learn generic features and dense layers extract domain specific features. However, this assumption has not been verified on the NILM context, thus it is unreliable to intuitively apply this presumption and its corresponding transfer approach to Seq2Point model. }

Based on the procedure given in Sec.~\ref{subsec:transfer_analysis}, we train a Seq2Point model to disaggregate the mains signal for a dish washer on REFIT dataset, and then transfer this model to detect the dish washer in REDD dataset. More specifically, we gradually transfer the seven layers in Seq2Point model, by either fixing or fine-tuning them and initializing the rest layers. The experimental results are shown in Fig.~\ref{fig5}, based on which we have the following observations and empirical findings.

\begin{itemize}
\item First, we observe the significantly increasing of both MAE and SAE along with the decreasing of F1-score, compared with the nearly steady performance from \emph{conv1} to \emph{conv2}, once we begin to share the fully connected layers. This implies that the convolutional layers virtually learn the generic features that are common in source and target domains, while the fully connected layers focus on extracting domain specific signatures. In other words, the number of transferred convolutional layers generally have little, if any, influence on the final prediction, while the more the dense layers we transfer the less the accuracy we get on the target domain. \nop{Arguably, we could simply freeze the convolutional layers during transfer learning without worrying about accuracy reduction.}
\item Second, we find that the NILM performance of fine-tuning transferred layers is better than fixing them, with lower MAE and SAE and higher F1-score. This indicates that fine-tuning the last dense layers could further help us promote the NILM performance on target domain.
\item Finally, we also observe a slight performance improvement as we procedurally transfer layers from \emph{conv1} to \emph{conv5}. Note that if we choose to freeze or fine-tune one particular convolutional layer, the parameters in this convolutional layer are pre-trained in source domain. Thus when transferring to target domain, the transferred layers could also leverage the information pre-trained in source domain, rather than learning from scratch. Arguably, the extra information from source domain enables the transfer learning model to generalize better on target domain.
\end{itemize}

\subsubsection{Effectiveness of Transfer Learning}

To validate the effectiveness of local transfer learning, particularly the CORAL approach, we conduct experiments on the edge devicxe of Nvidia Jetson Nano to simulate the procedure of local updates. Basically, we train a pruned MTL-Seq2Point model with one-month REDD data (for four appliances: washing machine,fridge, dish washer and microwave), which serves as the cloud model trained on source domain. Then, to validate the effectiveness of transfer learning, we leverage the pre-trained cloud model to detect appliances in UK-DALE dataset. Arguably, the REDD is literally distinct from the UK-DALE, as the households in former dataset are located in the US while those in latter dataset are sited in the UK. 

For the test without applying transfer learning, we directly employ the pre-trained cloud model on UK-DALE dataset for energy disaggregation. For the test with transfer learning, we first retrain the model using the deep CORAL approach and then leverage the retrained model to detect appliances in UK-DALE dataset. With results from the above two tests, we can compare their performances and see whether transfer learning takes effect. 

The results are shown in Table~\ref{table3}, from which we could observe that: for both model with sequence length 99 and 499, the model with local transfer learning outperforms the original one by up to 45\% in MAE, 85\% in SAE and 40\% in F1-score, respectively. In particular, for MAE, the performance is improved for all four appliances with transfer learning; for SAE, the accuracy is improved for three appliances; for F1-score, transfer learning model shows better performances on dish washer and microwave and comparable performance on fridge. In general, local transfer learning takes effect on minimizing the difference in source and target domains, compensating for the performance degradation due to domain shift.

In our implementations on the edge device, the model training time for each appliance is around 100 seconds in average, whereas the inference merely takes less than one second. During each local updating process, once receiving the cloud model, the edge client could perform transfer learning based on its own power readings to fine-tune the model parameters in dense layers, and conduct energy disaggregation with this personalized model promptly. Meanwhile, the updated local model would be uploaded to the cloud for further federated averaging, with the goal to collaboratively train more powerful and up-to-date cloud models.

\begin{table}[t]
\label{table3}
\caption{The effectiveness of transfer learning}
\begin{tabular}{|c|c|cc|cc|c|}
\hline
\multirow{2}{*}{Appliance}                                                 & \multirow{2}{*}{Metric} & \multicolumn{2}{c|}{\begin{tabular}[c]{@{}c@{}}None\\ Transfer\end{tabular}} & \multicolumn{2}{c|}{\begin{tabular}[c]{@{}c@{}}Transfer\\ Learning\end{tabular}} & \multirow{2}{*}{\begin{tabular}[c]{@{}c@{}}Average\\ Improvement\end{tabular}} \\ \cline{3-6}
                                                                            &                          & 99                                    & 499                                  & 99                                      & 499                                    &                                                                                \\ \hline
\multirow{3}{*}{\begin{tabular}[c]{@{}c@{}}Washing \\ Machine\end{tabular}} & MAE                      & 38.83                                 & 51.59                                & 25.43                                   & 23.54                                  & 45.84\%                                                                        \\
                                                                            & SAE                      & 7.79                                  & 1.58                                 & 0.87                                    & 0.78                                   & 82.39\%                                                                        \\
                                                                            & F1-score                       & 0.47                                  & 0.30                                 & 0.25                                    & 0.31                                   & -27.27\%                                                                       \\ \hline
\multirow{3}{*}{Fridge}                                                     & MAE                      & 68.78                                 & 68.75                                & 55.67                                   & 66.95                                  & 10.84\%                                                                        \\
                                                                            & SAE                      & 11.40                                 & 11.15                                & 0.88                                    & 2.36                                   & 86.63\%                                                                        \\
                                                                            & F1-score                       & 0.58                                  & 0.58                                 & 0.58                                    & 0.58                                   & 0                                                                              \\ \hline
\multirow{3}{*}{\begin{tabular}[c]{@{}c@{}}Dish\\ Washer\end{tabular}}      & MAE                      & 13.54                                 & 14.21                                & 11.18                                   & 10.29                                  & 22.63\%                                                                        \\
                                                                            & SAE                      & 1.17                                  & 0.76                                 & 2.21                                    & 2.49                                   & -143.52\%                                                                      \\
                                                                            & F1-score                       & 0.05                                  & 0.18                                 & 0.02                                    & 0.31                                   & 43.48\%                                                                        \\ \hline
\multirow{3}{*}{Microwave}                                                  & MAE                      & 41.87                                 & 31.85                                & 25.21                                   & 23.76                                  & 33.57\%                                                                        \\
                                                                            & SAE                      & 3.01                                  & 2.17                                 & 2.08                                    & 1.97                                   & 21.81\%                                                                        \\
                                                                            & F1-score                       & 0.30                                  & 0.15                                 & 0.31                                    & 0.32                                   & 40.00\%                                                                        \\ \hline
\end{tabular}
\end{table}

\balance 

\section{Conclusion}

We presented FedNILM, a federated learning paradigm designed for privacy-preserving and personalized NILM applications at low-end edge devices. FedNILM realized data privacy-preserving through federated learning, efficient model compression via filter pruning and multi-task learning, and personalized model building by unsupervised transfer learning, respectively. The results from the experiments on real-world energy data demonstrate that, FedNILM can achieve accurate and personalized energy disaggregation without compromising the user privacy.

\bibliographystyle{IEEEtran}  
\bibliography{references}  

\nop{\begin{IEEEbiography}{Michael Shell}
Biography text here.
\end{IEEEbiography}

\begin{IEEEbiographynophoto}{John Doe}
Biography text here.
\end{IEEEbiographynophoto}


\begin{IEEEbiographynophoto}{Jane Doe}
Biography text here.
\end{IEEEbiographynophoto}}

\end{document}